\documentclass[hyphens]{article}

\usepackage{microtype}
\usepackage{graphicx}

\usepackage{booktabs} 

\usepackage{hyperref}

\usepackage[accepted]{icml2025}
\usepackage{amsmath}
\usepackage{amssymb}
\usepackage{mathtools}
\usepackage{amsthm}

\usepackage[capitalize,noabbrev]{cleveref}

\theoremstyle{plain}

\theoremstyle{definition}

\theoremstyle{remark}

\usepackage[textsize=tiny]{todonotes}

\usepackage{subcaption}

\icmltitlerunning{High-Resolution Live Fuel Moisture Content (LFMC) Maps for Wildfire Risk from Multimodal Earth Observation Data}

\begin{document}

\twocolumn[
\icmltitle{High-Resolution Live Fuel Moisture Content (LFMC) Maps \\ for Wildfire Risk from Multimodal Earth Observation Data}

\begin{icmlauthorlist}
\icmlauthor{Patrick Alan Johnson}{ai2}
\icmlauthor{Gabriel Tseng}{ai2,mila,mcgill}
\icmlauthor{Yawen Zhang}{ai2}
\icmlauthor{Heather Heward}{ui}
\icmlauthor{Virginia Sjahli}{ai2}
\icmlauthor{Favyen Bastani}{ai2}
\icmlauthor{Joseph Redmon}{ai2}
\icmlauthor{Patrick Beukema}{ai2}

\end{icmlauthorlist}

\icmlaffiliation{mila}{Mila -- Quebec AI Institute}
\icmlaffiliation{mcgill}{McGill University}
\icmlaffiliation{ai2}{Allen Institute for AI (Ai2)}
\icmlaffiliation{ui}{University of Idaho}

\icmlcorrespondingauthor{Patrick Alan Johnson}{patrickj@allenai.org}

\icmlkeywords{Machine Learning, ICML}

\vskip 0.3in
]

\printAffiliationsAndNotice{}  

\begin{abstract}
Wildfires are increasing in intensity and severity at an alarming rate. Recent advances in AI and publicly available satellite data enable monitoring critical wildfire risk factors globally, at high resolution and low latency. Live Fuel Moisture Content (LFMC) is a critical wildfire risk factor and is valuable for both wildfire research and operational response. However, ground-based LFMC samples are both labor intensive and costly to acquire resulting in sparse and infrequent updates. In this work, we explore the use of a pretrained, highly-multimodal earth-observation model for generating large-scale spatially complete (wall-to-wall) LFMC maps. Our approach achieves significant improvements over previous methods using randomly initialized models ($>20\%$ reduction in RMSE). We provide an automated pipeline that enables rapid generation of these LFMC maps across the United States, and demonstrate its effectiveness in two regions recently impacted by wildfire (Eaton and Palisades).
\end{abstract}

\section{Introduction}
\label{introduction}

Live Fuel Moisture Content (LFMC) is a measurement of the amount of water in live vegetation. It is a critical parameter for both wildfire research and operations. LFMC has been shown to have a strong influence on wildfire ignition \citep{chuvieco2004conversion}, fuel availability \citep{kelley2019contemporary}, and wildfire spread \citep{rossa2017effect}. Live Fuel Moisture Content is calculated using the following formula:

\begin{equation}
    \mathrm{LFMC}\ \lbrack\%\rbrack\  = \ \frac{W_f\  - \ W_d}{W_d}\  \times 100
\end{equation}

Where $W_f$ is the weight of fresh plant material and $W_d$ is the weight of the material after drying (often in an oven). Lower LFMC values indicate higher wildfire risk \cite{rao2023dry}.

LFMC sampling in the field is a time-consuming and labor-intensive process. Variations in land cover, plant species, and topography can all significantly impact LFMC.  Effective and accurate sampling requires expert judgment. Plant material is collected, weighed, dried for one to two days in an oven, and weighed again to calculate LFMC as the percentage change in weight. Depending on site access, equipment, and fuel type, sampling just a single site takes 12 hours -- 4 days. Due to these constraints, only a few samples can be gathered at a time, limiting spatial coverage. 

Accurate, \emph{spatially complete}, assessments of LFMC have broad applications, including historical analysis of wildfires, planning prescribed burns, and supporting decision-making during active wildfires. Recent work has investigated a combination of machine learning and remote sensing-based techniques to provide wall-to-wall LFMC maps \cite{miller2023projecting,rao2020sar,marino2020investigating}. In tandem, the machine learning community has been investigating self-supervised models, with the stated intent of improving the spatial and temporal generalization of machine learning models applied to remote sensing \cite{astruc2024anysat,szwarcman2024prithvi,tseng2023lightweight}. These models can be critical tools for LFMC predictions, which often suffer from imbalanced and heterogeneous labels.

In this work, we develop a pipeline which leverages Galileo \cite{tseng2025galileo} --- a pretrained remote sensing model --- to generate on-demand LFMC maps. We validate the effectiveness of using a pretrained model, and demonstrate the effectiveness of our pipeline in two study areas.

The contributions of this paper are threefold:
\begin{itemize}
\item We develop a pipeline to generate LFMC maps for spatiotemporal areas of interest, allowing users to generate LFMC maps for future treatment planning and in response to historical wildfire events.
\item We demonstrate the utility of a highly multimodal \emph{pretrained} geospatial model to make accurate LFMC predictions.
\item We apply our pipeline to two real-world case studies without available labels, showing that the model’s predictions of LFMC are consistent with expert field observations.
\end{itemize}

The code is available on GitHub~\footnote{\url{https://github.com/allenai/lfmc}}.

\section{Related Works}

Previous work on estimating LFMC with machine learning has primarily focused on leveraging multimodal remote sensing data and \emph{fully supervised} approaches to generate coarse resolution maps. For example, \citet{rao2020sar} developed a physics-assisted, recurrent neural network (RNN) model to map the LFMC for the western United States at a 250-meter spatial resolution, mainly leveraging Sentinel-1 backscatter and Landsat-8 optical imagery. \citet{miller2023projecting} proposed a temporal convolutional neural network (tempCNN) model to forecast LFMC at a 500-meter resolution with a three-month lead time, mainly using ground observations, MODIS images, meteorological variables, climate zone classifications, and elevation data. In this study,  we investigate a model \emph{pretrained} on multimodal remote sensing data, and produce maps at a far higher spatial resolution (10-meter). 

As part of efforts to modernize the U.S. National Fire Danger Rating System, \citet{jolly2024modernizing} introduced a physiologically based LFMC model grounded in plant response mechanisms. Their approach uses a Growing Season Index (GSI) model primarily driven by daily surface weather variables, including minimum temperature, vapor pressure deficit, photoperiod, and rainfall derived from both point-source and gridded weather observations and forecasts. In contrast, we investigate an empirical model that relies on observation data, without explicitly modeling plant physiological processes.

\section{Materials and methods}
\label{materials-and-methods}

\subsection{The Globe-LFMC 2.0 dataset}
\label{lfmc-samples-dataset}

In 2024, the second iteration of the global LFMC dataset was published, Globe-LFMC 2.0 \citep{yebra2024globelfmc}. This dataset comprises over 280,000 LFMC values derived from samples gathered at more than 2,000 locations across 15 countries collected between 1977 and 2023. It includes data from more than 500 different species or combinations of species. 

Filtering and aggregating the Globe-LFMC 2.0 dataset resulted in 41,214 samples across 1,031 sites (described in Section \ref{sec:data-preparation}). The median LFMC value was 102.0\%, the 1\% LFMC value was 51\%, and the 99.9\% LFMC value was 301.67\%. To prevent outliers from having an outsized impact on predictions, all samples were capped and normalized with an LFMC value of 302\%, rounded up from the 99.9 percentile.

\begin{figure}[ht]
    \centering
    \includegraphics[width=0.9\linewidth]{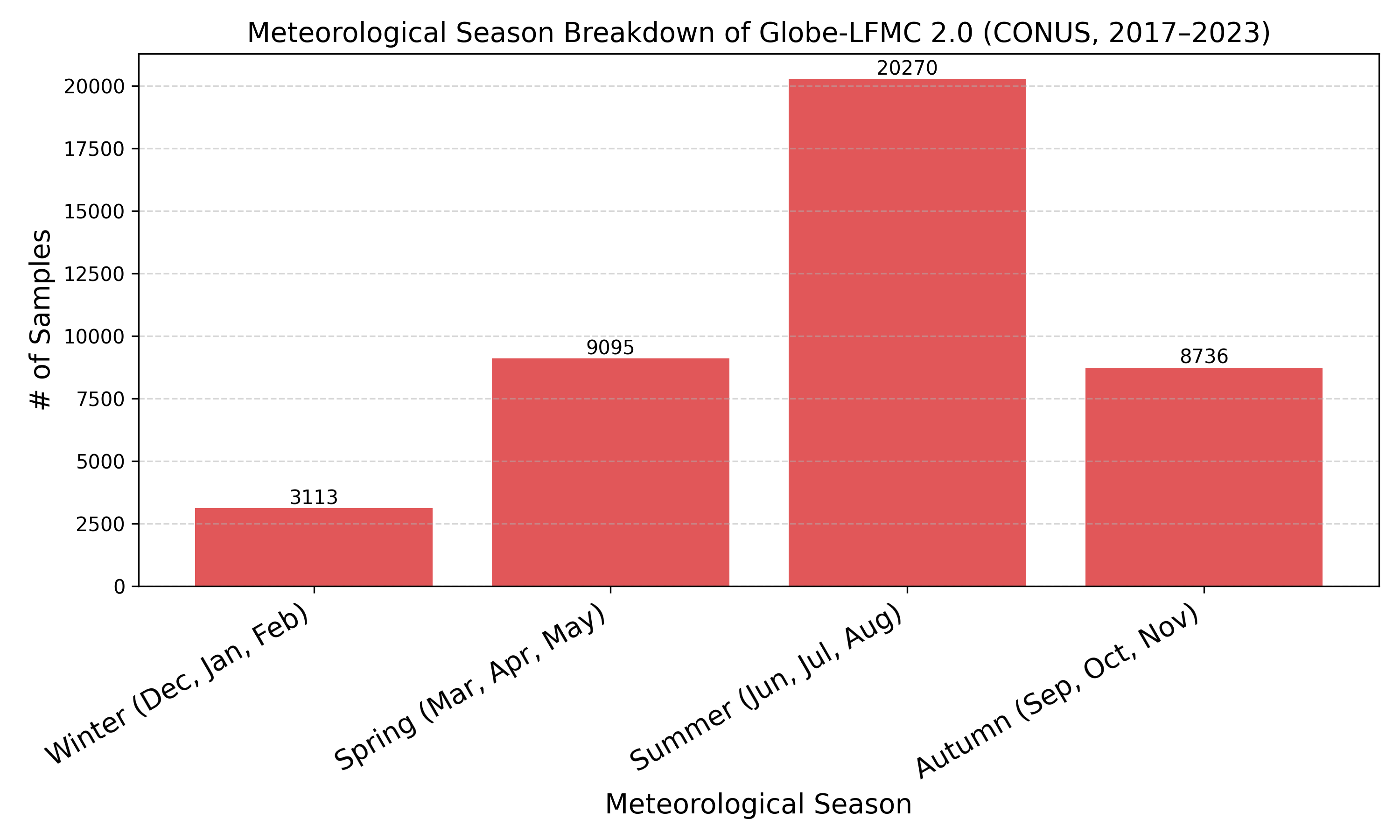}
    \caption{Meteorological season breakdown of the Globe-LFMC 2.0 dataset for CONUS samples 2017-2023.}
    \label{fig:lfmc-season-breakdown}
\end{figure}

As shown in Figure \ref{fig:lfmc-season-breakdown}, when grouped by meteorological season, most samples were taken during summer and the fewest were taken during winter. Spring and autumn had a similar number of samples. This aligns with the typical annual wildfire risk during a calendar year in the United States. Additionally, the spring and autumn months are often used for forest treatments such as prescribed burning due to more favorable weather conditions \citep{hosten2020fireweather}.

\begin{figure*}[htbp]
  \centering
  \begin{minipage}{0.75\textwidth}
    \centering
    \includegraphics[width=\linewidth]{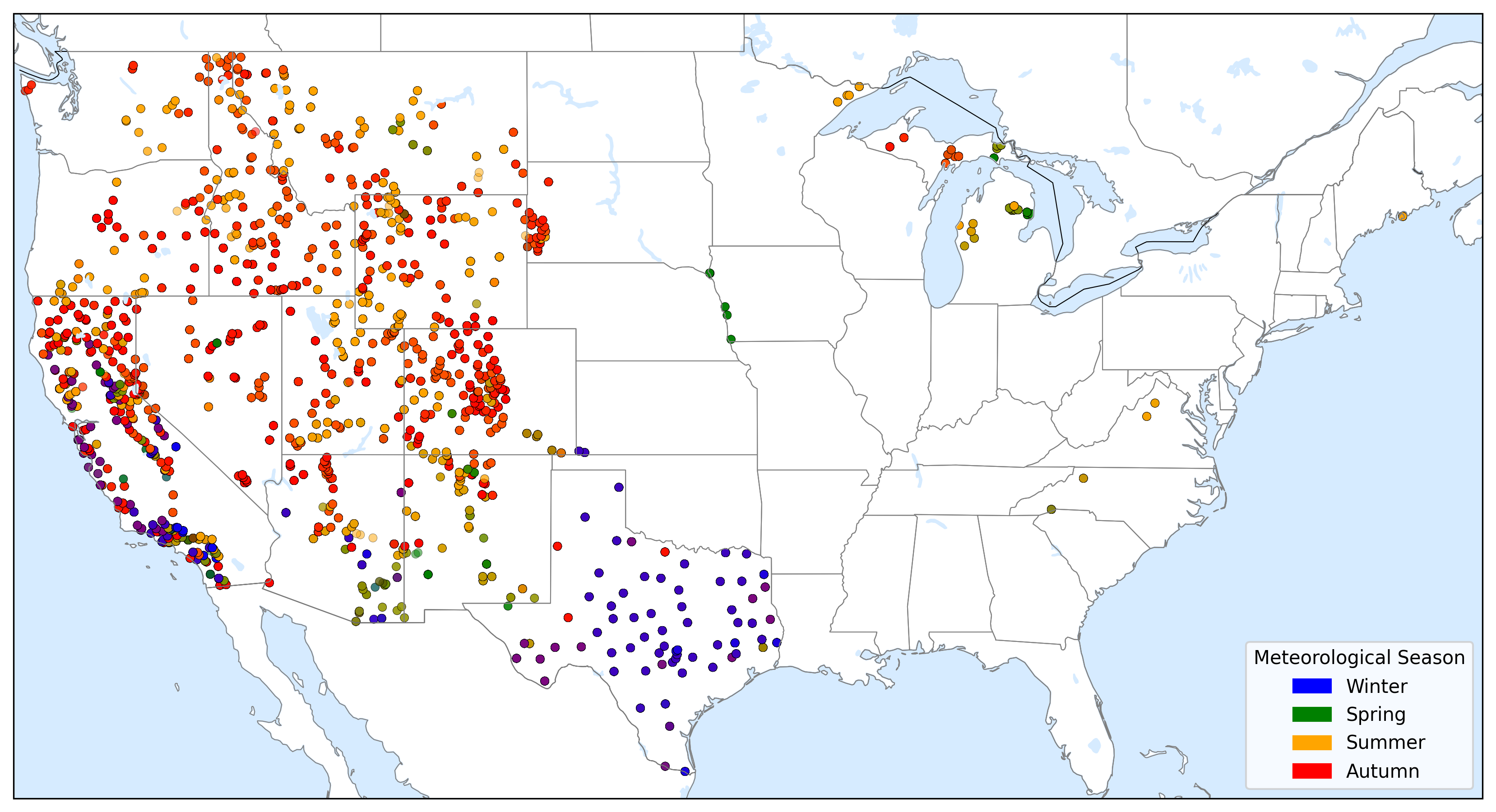}
    \caption{Location breakdown of Globe-LFMC 2.0 dataset for CONUS samples 2017-2023 by meteorological season.}
    \label{fig:location-samples}
  \end{minipage}
\end{figure*}

Also noticeably, as shown in Figure \ref{fig:location-samples}, most winter samples were taken in warmer climates such as Southern California, Arizona, and Texas.

\subsection{The Galileo Pretrained Remote Sensing Model}

Recent advancements have created new opportunities for applications that combine remote sensing with deep learning \citep{vatsavai2024geospatial}. Notably, by pretraining on a large volume of modalities across time and space, geospatial foundation models enable fine-tuned models to be produced efficiently and with lower data labeling requirements than traditional machine learning techniques \citep{jakubik2023foundation}. A well trained foundation model can be adapted to a diverse set of remote sensing applications, from crop mapping to land cover classification to flood detection \citep{dionelis2024evaluating}.

LFMC estimation is a highly multimodal task; \citet{rao2020sar} report the beneficial impact of including a diversity of remote sensing modalities, ranging from directly sensed products such as optical and SAR data to derived products including topography and land cover. We therefore require a \emph{highly multimodal} foundation model, which can process this diversity of inputs. We select the Galileo model \cite{tseng2025galileo}, a model which can ingest 10 directly-sensed and derived remote sensing products across both the spatial and temporal dimensions. We document the inputs we used in more detail below.

Galileo is a vision transformer based model \cite{dosovitskiy2020image} --- we use a Galileo-Tiny model, which contains 5.3M parameters. The Galileo-Tiny model is chosen for its balance between a lightweight architecture and strong performance.

\subsubsection{Galileo's remote sensing inputs}
\label{sec:remote-sensing-data}

We use a subset of Galileo's remote sensing inputs when producing our LFMC maps. All remote sensing data was collected from the Google Earth Engine Data Catalog \citep{gorelick2017googleee}.

\begin{itemize}
\item \textbf{Multispectral Optical Data}: We use optical imagery from the Sentinel-2 satellite, which include the visible, near-infrared and shortwave infrared bands. In addition, we compute NDVI \cite{tucker1979red} from the near-infrared and red bands. Following Galileo's pretraining, we use the L1C Top of Atmosphere (TOA) product.
\item \textbf{Synthetic Aperture Radar Data}: We include the dual VV and VH bands from the Sentinel-1 satellite, which can be useful for distinguishing between water, land and vegetation.
\item \textbf{Night Lights}: We use the night lights from the Visible Infrared Imaging Radiometer Suite (VIIRS) missions, including the average (Day/Night Band) DNB radiance values.
\item \textbf{Weather Data}: We include precipitation and temperature data from the ERA-5 Land reanalysis product \cite{munozsabater2019era5land}.
\item \textbf{Climate and Water Balance Data}: We include climate water deficit, soil moisture, and actual evapotranspiration from the TerraClimate dataset \cite{abatzoglou2018terraclimate}.
\item \textbf{Topography Data}: We include elevation and slope computed from the SRTM Digital Elevation Model \cite{farr2000shuttle}.
\item \textbf{Location Data} We provide location awareness to the model via latitude and longitude coordinates.
\end{itemize}

The input datasets used in this analysis span a wide range of spatial and temporal resolutions—for example, spatial resolution ranges from approximately 10 meters per pixel for certain Sentinel-2 bands to tens of kilometers per pixel for ERA5 weather data. Temporal resolution also varies considerably, from a 5-day revisit period for Sentinel-2 to monthly intervals for datasets such as ERA5, TerraClimate, and VIIRS. Some inputs, such as SRTM, are static and do not vary over time. To accommodate these differences, Galileo categorizes inputs based on whether they vary spatially, temporally, or both. Inputs with coarse spatial resolution, such as weather data, are treated as spatially static within the context of a single prediction instance. All temporally varying inputs are aggregated or resampled to a consistent monthly time scale to ensure comparability across datasets.

\subsection{Full coverage LFMC mapping pipeline}

\begin{figure}[ht]
\vskip 0.2in
\begin{center}
\centerline{\includegraphics[width=\columnwidth]{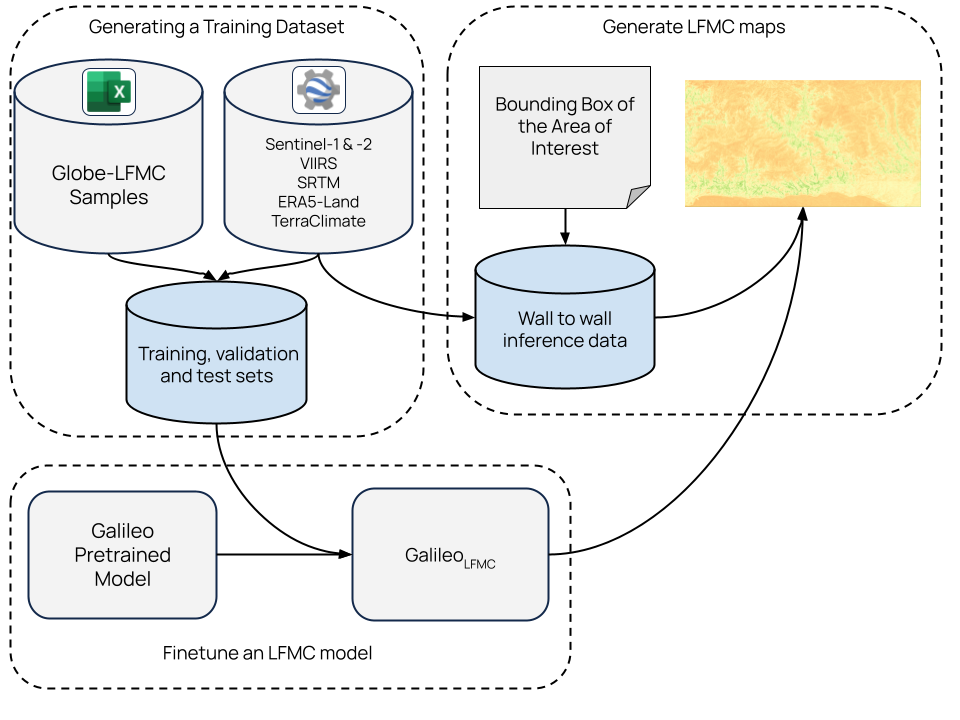}}
\caption{A pipeline to produce LFMC map for a given spatiotemporal window. By leveraging a pretrained remote sensing model and the Globe-LFMC 2.0 dataset, we can efficiently generate maps for new areas and timesteps.}
\label{fig:pipeline}
\end{center}
\vskip -0.2in
\end{figure}

We combine the Globe-LFMC dataset and the pretrained Galileo model to generate a fine-tuned LFMC model. We then apply this fine-tuned model to generate wall-to-wall LFMC maps. We outline this pipeline in Figure \ref{fig:pipeline}.

\subsubsection{Generating a training dataset from Globe LFMC-2.0}
\label{sec:data-preparation}

The Globe-LFMC dataset includes samples from around the globe, though the vast majority originate from the continental United States (CONUS). Accordingly, this study focuses solely on samples within CONUS. To ensure satellite data availability for all data sources, only samples from 2017 to 2023 were included.

In the Globe-LFMC 2.0 dataset, it is common to find multiple samples collected at the same location on the same day. Following the guidance of the Globe-LFMC 2.0 paper, such samples were averaged into a single observation.

For each sample in the Globe-LFMC 2.0 dataset, a 1 km $\times$ 1 km bounding box was created around its latitude--longitude coordinates. Data within this bounding box was exported from Google Earth Engine for the following products: Sentinel-1, Sentinel-2, VIIRS, SRTM, ERA5-Land, and TerraClimate, and elevation data from the SRTM product.

\begin{figure}[ht]
    \centering
    \includegraphics[width=0.9\linewidth]{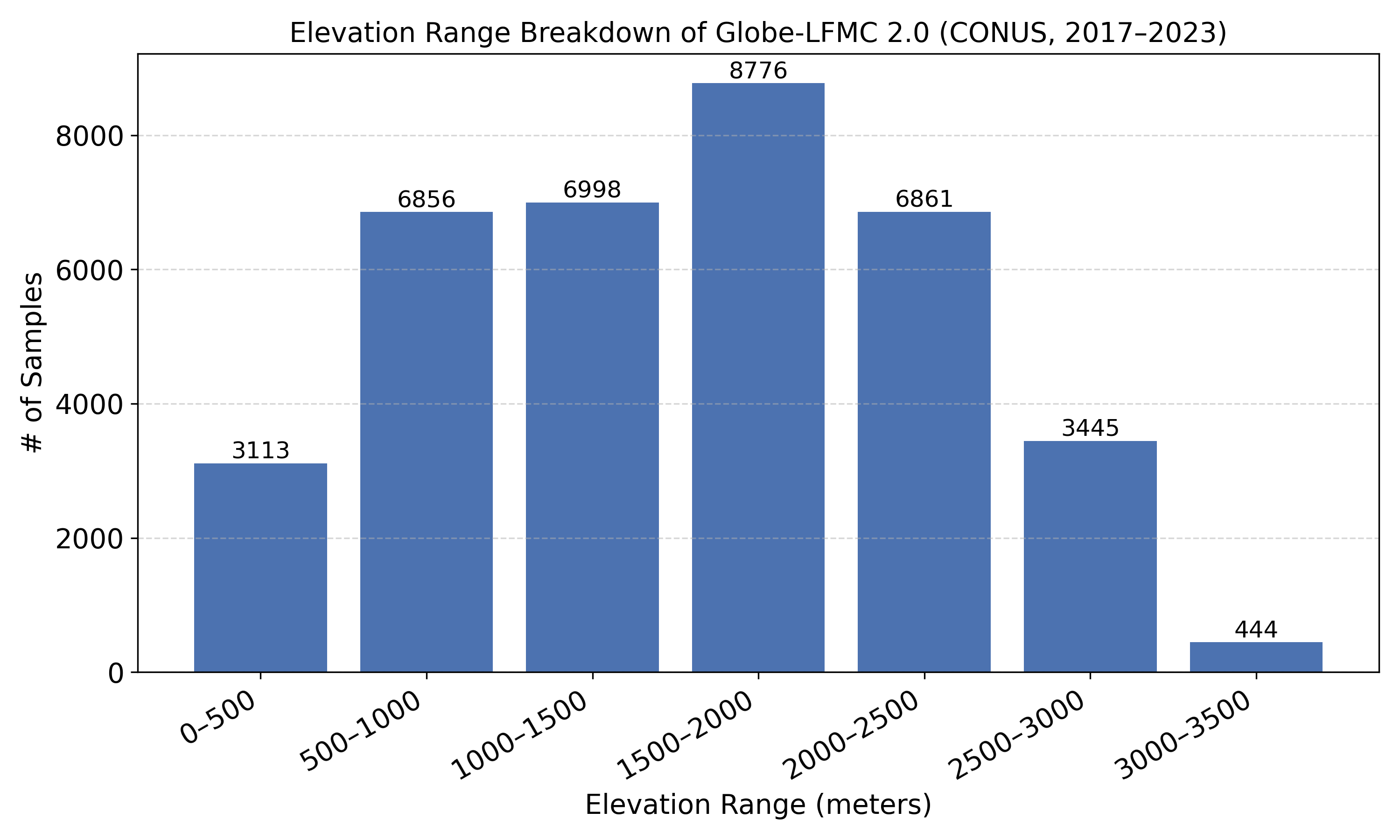}
    \caption{Elevation range breakdown of the Globe-LFMC 2.0 dataset for CONUS samples (2017–2023) in 500\,m increments.}
    \label{fig:lfmc-elevation-breakdown}
\end{figure}

\begin{figure}[ht]
    \centering
    \includegraphics[width=0.9\linewidth]{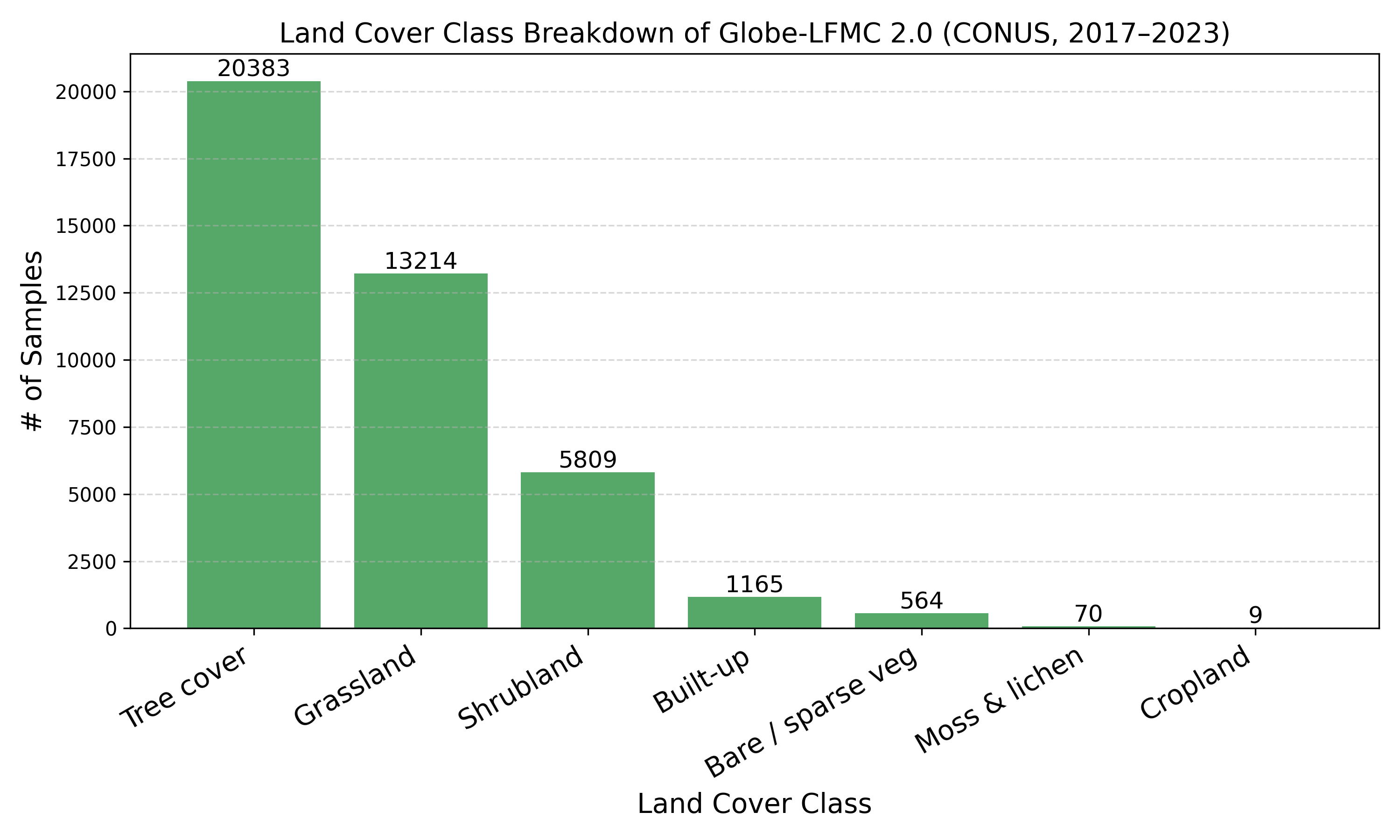}
    \caption{Coverage of the Globe-LFMC 2.0 dataset for CONUS samples 2017-2023 in terms of land cover classes covered by the labels. We use WorldCover \cite{zanaga2021esa} to obtain the land cover labels.}
    \label{fig:lfmc-lc-breakdown}
\end{figure}

In total, the dataset included 41,214 ground-based LFMC samples, randomly split into training ($\approx$ 70\%), validation ($\approx$ 15\%), and test ($\approx$ 15\%) sets. Sample elevations ranged from 15 to 3,187 meters and were grouped into 500-meter ranges for analysis, as detailed in Figure \ref{fig:lfmc-elevation-breakdown}. The breakdown of all land cover classifications (defined by the ESA WorldCover 2021 product) is shown in Figure \ref{fig:lfmc-lc-breakdown}.

\subsubsection{Fine-tuning an LFMC Model}
\label{sec:finetuning}

We fine-tuned the Galileo-Tiny pretrained on the Globe LFMC-2.0 dataset (combined with the exported remote sensing data, described above in Section \ref{sec:data-preparation}) under mean squared error (MSE) loss.

During fine-tuning, we used the $\approx$70\% training set and used the $\approx$15\% validation set as an early stopping mechanism. Training was conducted for up to 100 epochs, with early stopping applied if no improvement was observed against the validation set for 5 consecutive epochs. This yielded a fine-tuned Galileo$_{LFMC}$ model, which could then be deployed to generate LFMC maps.

\subsubsection{Generating LFMC maps}

Generating LFMC maps requires a fine-tuned Galileo$_{LFMC}$ model, and a bounding box describing the desired mapping area. We export the relevant remote sensing products using Google Earth Engine, and run inference over the exported area using Galileo$_{LFMC}$.

\subsubsection{Infrastructure}

The model was fine-tuned on a single NVIDIA H100 GPU.
Fine-tuning took approximately 30-60 minutes depending on the number of epochs it took for the model to reach a steady validation loss.

\section{Experiments and Results}

We report results on the (unseen) $\approx$ 15\% test set, using the random split described in Section \ref{sec:data-preparation}. We report root mean squared error (RMSE), mean absolute error (MAE) and the R\textsuperscript{2} score (coefficient of determination) of our predictions against the test set.

We compare the performance of Galileo$_{LFMC}$ against two baselines: (1) a fine-tuned model initialized with random weights and (2) a model that predicts values using the monthly average of all training samples corresponding to the sample's month. Previous works applying deep learning to LFMC estimation have adopted a fully supervised approach \cite{rao2020sar,miller2023projecting,zhu2021live} - our ``random weights'' baseline reflects this previous approach.

\subsection{Results}
\label{results}

\begin{table}[t]
\caption{Using a pretrained model significantly improves performance compared to a randomly initialized model. In this table, we compare a pretrained Galileo architecture to two baselines: (1) a randomly initialized model and (2) a monthly average, and find a significant reduction in RMSE and an increase in $R^2$ score.}
\label{tab:overall_results}
\vskip 0.15in
\begin{center}
\begin{small}
\begin{sc}
\begin{tabular}{lcccr}
\toprule
Category & RMSE & MAE & R\textsuperscript{2} \\
\midrule
Pretrained & \textbf{18.91} & \textbf{12.58} & \textbf{0.72} \\
Random initialized & 23.61 & 16.33 & 0.57 \\
Monthly predictions & 33.66 & 25.38 & 0.11 \\
\bottomrule
\end{tabular}
\end{sc}
\end{small}
\end{center}
\vskip -0.1in
\end{table}

\subsubsection{Pretrained weights achieve the best results}

Galileo$_{LFMC}$ achieved an overall MAE of 12.58 and an RMSE of 18.91, with an R\textsuperscript{2} score of 0.72 (Table \ref{tab:overall_results}). This is a significant improvement over the randomly initialized model ($\approx$ 20\% reduction in RMSE), highlighting the effectiveness of pretrained models for LFMC predictions. Our results are comparable to, and in some cases exceed, the performance of existing LFMC models developed without the use of pretrained weights \cite{miller2023projecting,rao2020sar,marino2020investigating}.

\begin{table}[t]
\caption{Fine-tuned model evaluation across meteorological seasons. Galileo$_{LFMC}$ performs well across all seasons - including the winter season, where it sees fewer training data points.}
\label{tab:fine-tuned-results_season}
\vskip 0.15in
\begin{center}
\begin{small}
\begin{sc}
\begin{tabular}{lcccr}
\toprule
Season & RMSE & MAE & R\textsuperscript{2} \\
\midrule
Overall & 18.91 & 12.58 & 0.72 \\
Winter season & 15.31 & 10.74 & 0.77 \\
Spring season & 22.85 & 15.35 & 0.69 \\
Summer season & 19.70 & 13.05 & 0.67 \\
Autumn season & 12.70 & 9.27 & 0.75 \\
\bottomrule
\end{tabular}
\end{sc}
\end{small}
\end{center}
\vskip -0.1in
\end{table}

\begin{table}[t]
\caption{Fine-tuned model evaluation across land cover classes. Galileo$_{LFMC}$ performs well across all land cover types.}
\label{tab:fine-tuned-results_lc}
\vskip 0.15in
\begin{center}
\begin{small}
\begin{sc}
\begin{tabular}{lcccr}
\toprule
Land Cover Class & RMSE & MAE & R\textsuperscript{2} \\
\midrule
Overall & 18.91 & 12.58 & 0.72 \\
Trees & 18.00 & 11.97 & 0.68 \\
Grass & 20.09 & 13.62 & 0.73 \\
Shrub & 19.53 & 12.28 & 0.74 \\
Built-up & 16.79 & 11.78 & 0.77 \\
Bare / Sparse & 20.52 & 15.67 & 0.79 \\
\bottomrule
\end{tabular}
\end{sc}
\end{small}
\end{center}
\vskip -0.1in
\end{table}

\begin{table}[t]
\caption{Fine-tuned model evaluation across elevation bands. R$^2$ values drop at higher elevations, likely due to a sparsity of training labels above 3,000m.}
\label{tab:fine-tuned-results_elevation}
\vskip 0.15in
\begin{center}
\begin{small}
\begin{sc}
\begin{tabular}{lcccr}
\toprule
Category & RMSE & MAE & R\textsuperscript{2} \\
\midrule
Overall & 18.91 & 12.58 & 0.72 \\
Elevation: 0-500m & 18.34 & 11.59 & 0.73 \\
Elevation: 500-1000m & 17.93 & 11.98 & 0.77 \\
Elevation: 1000-1500m & 21.65 & 14.54 & 0.73 \\
Elevation: 1500-2000m & 18.91 & 13.56 & 0.75 \\
Elevation: 2000-2500m & 19.35 & 12.44 & 0.61 \\
Elevation: 2500-3000m & 15.25 & 10.00 & 0.54 \\
Elevation: 3000-3500m & 14.54 & 10.41 & 0.32 \\
\bottomrule
\end{tabular}
\end{sc}
\end{small}
\end{center}
\vskip -0.1in
\end{table}

\begin{figure*}[htbp]
  \centering
  \begin{minipage}{0.75\textwidth}
    \centering
    \includegraphics[width=\linewidth]{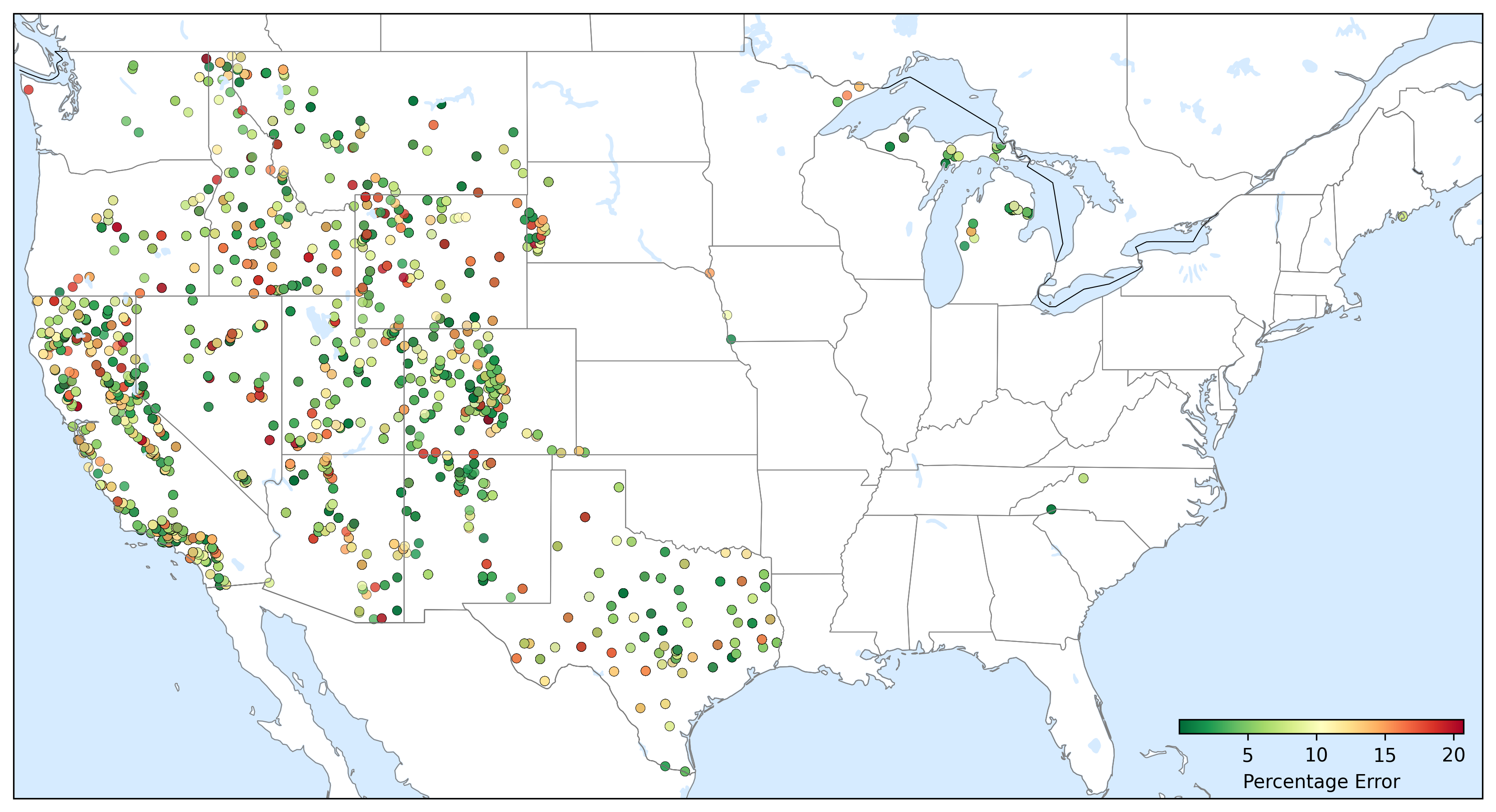}
    \caption{Percentage error for the test dataset. Galileo$_{LFMC}$ performs well across the entire spatial distribution of the test set. Moran’s I is 0.057 with a p-value of 0.001, indicating a statistically significant weak positive autocorrelation in model errors.}
    \label{fig:spatial_error}
  \end{minipage}
\end{figure*}

In addition to the results in Table \ref{tab:overall_results}, we plot model performance spatially in Figure \ref{fig:spatial_error}. To assess the spatial autocorrelation of the residuals, we compute a Moran's I \cite{moran1950notes} using K-Nearest Neighbors based spatial weights matrix. The results yielded a Moran\textquotesingle s I of 0.057 with a p-value of 0.001, suggesting a statistically significant weak positive spatial autocorrelation in the model errors. A weak but statistically significant spatial autocorrelation suggests potential information leakage from the random split, as geographically proximate samples may appear across training, validation, and test sets. Future work could adopt spatial partitioning methods to better assess model generalization.

When analyzed across meteorological seasons (Table \ref{tab:fine-tuned-results_season}), the model\textquotesingle s lowest RMSE was for autumn (12.70) and its highest RMSE was for spring (22.85). It did not demonstrate lower performance nor explainability (R\textsuperscript{2} score) in the winter, despite lower representation in the dataset. This suggests that Galileo's pretraining---which exposes it to data from many different timesteps---is helpful in ensuring Galileo$_{LFMC}$ can generalize temporally.

The model's performance across the various land cover classes was consistent with its overall accuracy (Table \ref{tab:fine-tuned-results_lc}). RMSE values ranged from 16.79 for built-up areas to 20.52 for sparse vegetation, with a median value of 18.00. Notably, despite their lower representation, built-up and sparse areas did not exhibit significantly different errors compared to more prevalent classes such as tree-covered areas, grasslands, and shrublands. Land cover types with very limited representation in the filtered Globe-LFMC 2.0 dataset---specifically moss/lichen and cropland---were not covered by the test set.

The model maintained an R\textsuperscript{2} score above 0.7 for elevation ranges below 2,000 meters (Table \ref{tab:fine-tuned-results_elevation}). However, performance declined substantially at higher elevations, with R\textsuperscript{2} values dropping to 0.54 for the 2,500--3,000 meter range and 0.32 for the 3,000--3,500 meter range. This decrease is likely due to the limited representation of high-elevation areas in the Globe-LFMC 2.0 dataset---in particular, there were only 444 samples above 3,000 meters. RMSE values across elevation bands ranged from 14.54 (3000--3500 meters) to 21.65 (1000--1500 meters). Among the bands with R\textsuperscript{2} \textgreater{} 0.7---those under 2,000 meters---the lowest RMSE was 17.93. The overall median RMSE was 18.34.

\subsubsection{Model performance is not sensitive to input shape}

In addition to being highly multimodal, Galileo can ingest a variety of input shapes (i.e. varying spatial areas and temporal ranges). We investigate the effect of passing different input shapes as inputs (compared to our default spatial area of $[32 \times 32]$ pixels and 12 timesteps) in Table \ref{tab:fine-tuned-results_shape}. While more spatial and temporal context helps, we find that the model remains relatively stable as the spatial area and temporal range are reduced.

\begin{table}[t]
\caption{We fine-tune the pretrained Galileo model with a variety of input shapes (i.e. varying [height$\times$ width] and number of input timesteps). While more spatial and temporal context is beneficial, the fine-tuned model's performance stays relatively stable as these parameters are changed.}
\label{tab:fine-tuned-results_shape}
\vskip 0.15in
\begin{center}
\begin{small}
\begin{sc}
\begin{tabular}{llcccr}
\toprule
H, W & T & RMSE & MAE & R\textsuperscript{2} \\
\midrule
32 & 12 & 18.91 & 12.58 & 0.72 \\
32 & 6 & 18.93 & 12.68 & 0.72 \\
32 & 3 & 19.45 & 13.10 & 0.70 \\
16 & 12 & 19.36 & 12.67 & 0.71 \\
8 & 12 & 19.71 & 13.20 & 0.70 \\
1 & 12 & 20.25 & 13.46 & 0.68 \\
\bottomrule
\end{tabular}
\end{sc}
\end{small}
\end{center}
\vskip -0.1in
\end{table}

\subsubsection{Pretraining makes the model more robust to missing inputs}

We use highly multimodal remote sensing inputs when fine-tuning the Galileo model and making LFMC maps (described in Section \ref{sec:remote-sensing-data}). To understand the contribution of each of these modalities, we fine-tune the pretrained and randomly initialized models with one input removed in Table \ref{tab:fine-tuned-results_inputs}. We find that for the randomly initialized model, removing TerraClimate yields a significant degradation in performance ({8\%} reduction in RMSE). On the other hand, the pretrained model's performance remains relatively stable even as inputs are removed, which suggests Galileo's self-supervised pretraining increases its robustness to missing inputs at fine-tuning time. This allows the pretrained model to be leveraged even if certain remote sensing products are missing (e.g. due to clouds). 

\begin{table}[t]
\caption{We fine-tune the pretrained and randomly initialized Galileo models while removing one of the at fine-tuning time. Pretraining makes the model more robust to missing inputs; the randomly initialized model experiences significantly larger changes in performance depending on the removed inputs (e.g. a 14 \% degradation in R$^2$ score when the TerraClimate input is removed).}
\label{tab:fine-tuned-results_inputs}
\vskip 0.15in
\begin{center}
\begin{small}
\begin{sc}
\begin{tabular}{l|ccc|ccc}
\toprule
& \multicolumn{3}{c|}{Pretrained} & \multicolumn{3}{c}{Random} \\ 
W/O IN & RMSE & MAE & R\textsuperscript{2} & RMSE & MAE & R\textsuperscript{2} \\
\midrule
None & \textbf{18.91} & 12.58 & 0.72 & 23.61 & 16.33 & 0.57 \\ 
S2 & 19.51 & 13.10 & 0.70 & 23.46 & 16.45 & 0.57 \\
S1 & \textbf{18.82} & 13.10 & 0.72 & 23.84 & 16.57 & 0.56 \\
ERA5 & 19.27 & 13.09 & 0.71 & \textbf{22.42} & 15.90 & 0.61 \\
TC & 19.51 & 13.02 & 0.70 & 25.57 & 17.59 & 0.49 \\
SRTM & 19.61 & 13.34 & 0.70 & \textbf{22.02} & 15.34 & 0.62 \\
loc. & 20.08 & 13.91 & 0.69 & 23.80 & 16.54 & 0.56 \\
\bottomrule
\end{tabular}
\end{sc}
\end{small}
\end{center}
\vskip -0.1in
\end{table}

\subsection{Case study: 2025 Palisades and Eaton Wildfires}\label{case-study-2025-palisades-and-eaton-wildfires}

In January 2025, a series of devastating wildfires burned across the Los Angeles metropolitan area. These fires were exacerbated by prolonged drought, low humidity, unusually strong northeasterly (Santa Ana) winds, and a buildup of vegetation following two consecutive springs of above-average rainfall \citep{digiuseppe2025global}.

The two largest and most destructive fires---by both area burned and structures lost---were the Palisades and Eaton wildfires, both located in Los Angeles County. The Palisades Fire originated in the Santa Monica Mountains and severely impacted Pacific Palisades, Topanga, and Malibu. The Eaton Fire started in the San Gabriel Mountains and destroyed foothill communities, particularly Altadena. Both fires caused unprecedented levels of damage.

Perimeter data for both fires was retrieved on January 13, 2025, from the WFIGS Current Interagency Fire Perimeters database \citep{nifc2025wfigs}. At the time, both fires were active and largely uncontained. Bounding boxes were constructed for each fire, and data was exported from Google Earth Engine covering the period from January 1, 2020, to December 31, 2024.

\begin{figure}
\centering
\includegraphics[width=\linewidth]{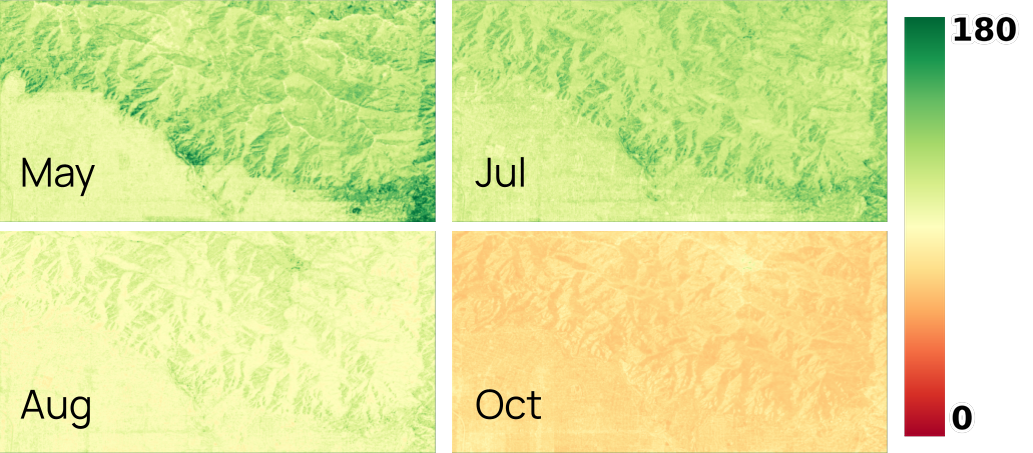}
\caption{LFMC predictions for the San Gabriel Mountains, site of the Eaton Fire in 2025. These maps show 4 months in 2024 and demonstrate our pipeline's ability to make spatially complete maps that capture seasonal patterns and geographic variability in LFMC values. LFMC values range from 0\% (red) to 180\% (green).}
\label{fig:eaton_2024}
\end{figure}

The fine-tuned model was used to generate monthly LFMC predictions for the affected areas, producing outputs at a 10-meter spatial resolution. We show a seasonal example of these predictions for the Eaton Fire in Figure~\ref{fig:eaton_2024}.

\begin{figure}[ht]
\centering
\begin{subfigure}[b]{0.95\linewidth}
    \centering
    \includegraphics[width=\linewidth]{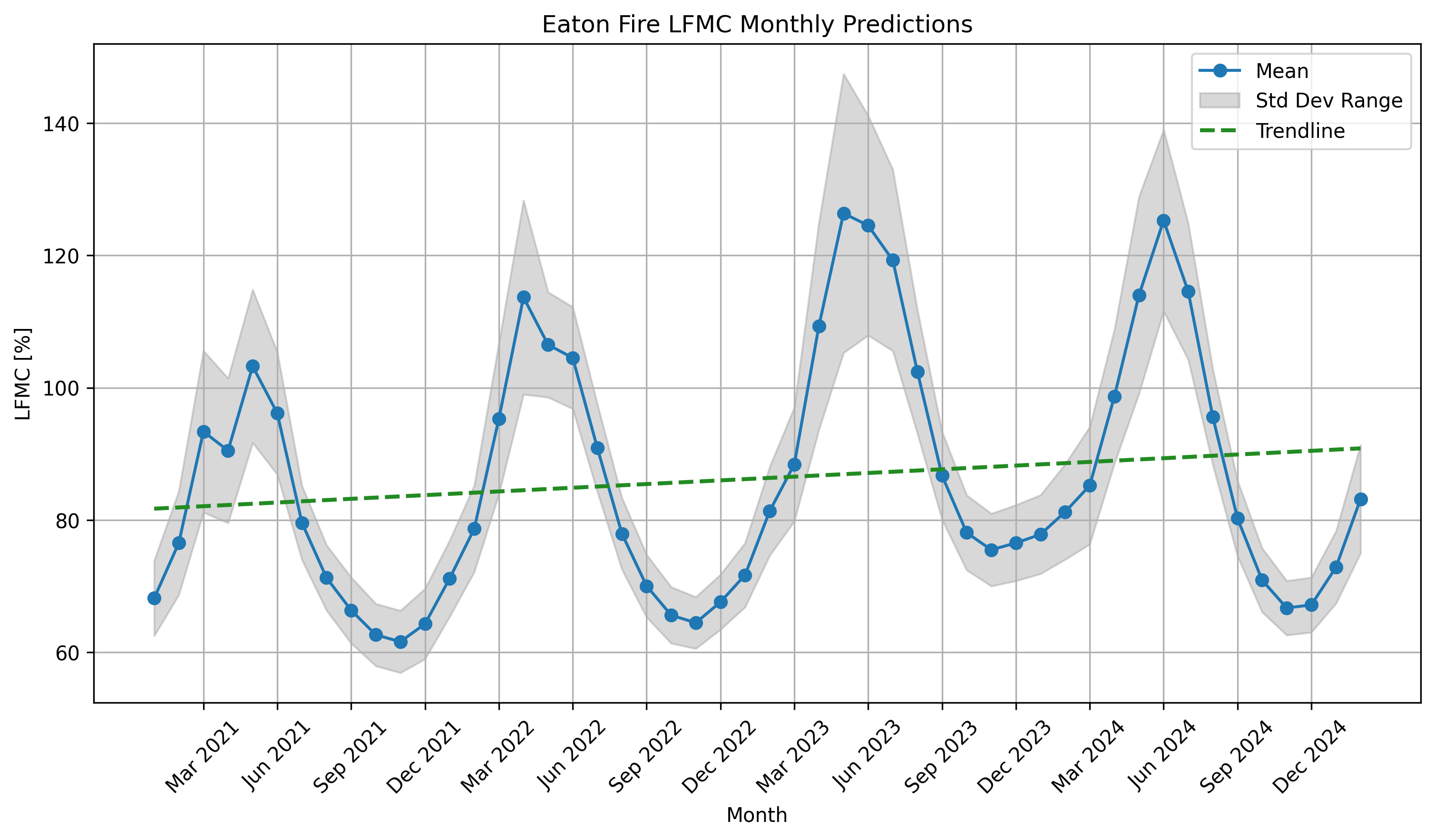}
    \caption{Eaton Fire area}
    \label{fig:eaton-monthly}
\end{subfigure}
\vskip 0.5em
\begin{subfigure}[b]{0.95\linewidth}
    \centering
    \includegraphics[width=\linewidth]{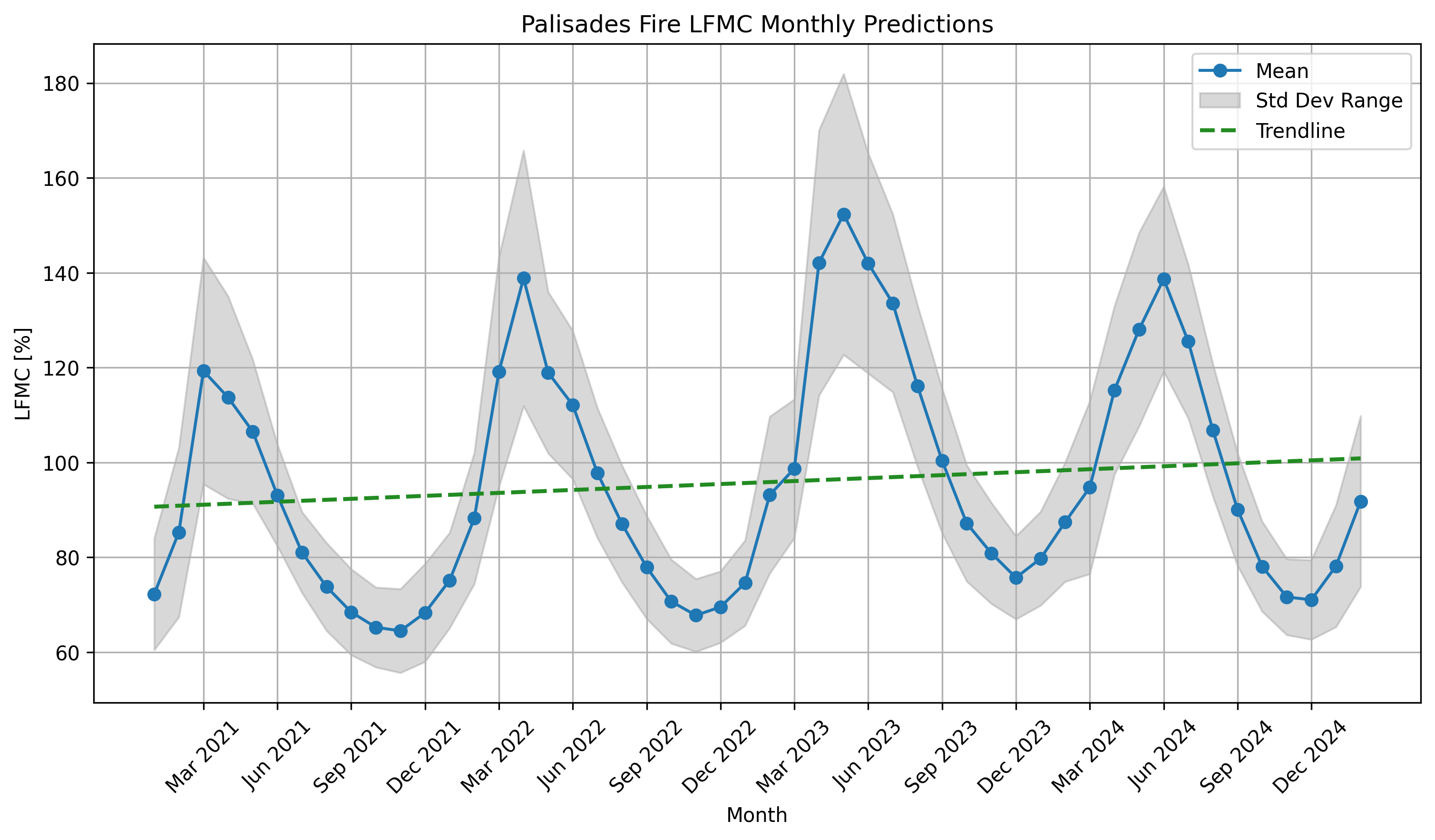}
    \caption{Palisades Fire area}
    \label{fig:palisades-monthly}
\end{subfigure}
\caption{Monthly LFMC average predictions for two wildfire areas from 2021–2024: (a) Eaton and (b) Palisades. These averages are computed over wall-to-wall maps like the ones shown in Figure~\ref{fig:eaton_2024} and Figure~\ref{fig:palisades_2024}.}
\label{fig:monthly-lfmc-comparison}
\end{figure}

\begin{figure}
\centering
\includegraphics[width=\linewidth]{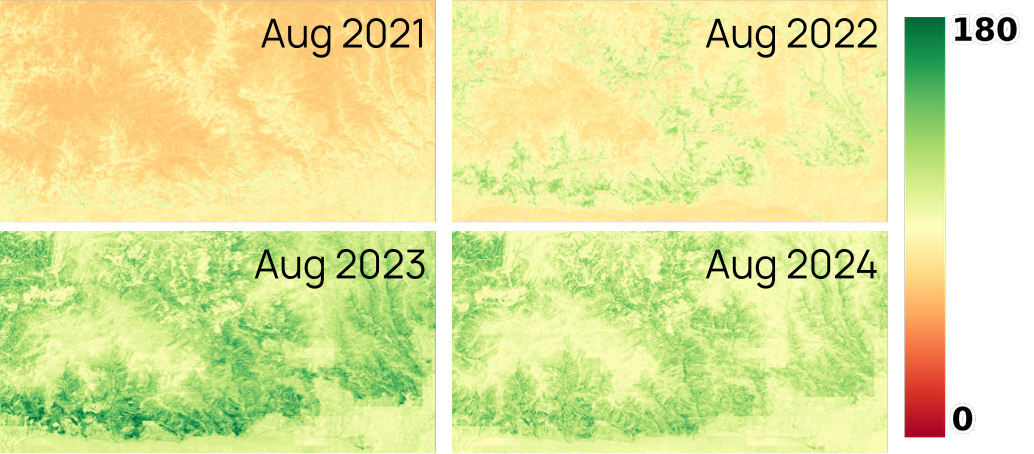}
\caption{LFMC predictions for the Santa Monica Mountains and nearby coastal communities of Pacific Palisades, Topanga, and Malibu, site of the Palisades fire in 2025. These maps show the month of August for 2021, 2022, 2023, and 2024, demonstrating our pipeline's ability to produce spatially complete maps that capture annual variability in LFMC values driven by differences in seasonal cycles. LFMC values range from 0\% (red) to 180\% (green).}
\label{fig:palisades_2024}
\end{figure}

The model revealed a consistent seasonal pattern: average LFMC levels rose through winter and peaked in late spring before declining into late autumn. This is shown in the average LFMC prediction for all pixels in the bounding boxes (see Figure~\ref{fig:monthly-lfmc-comparison}). Additionally, the model predicted higher LFMC values in 2023 and 2024 compared to 2021 and 2022 (Figure~\ref{fig:palisades_2024}). While the link between LFMC and fuel accumulation is not fully understood, this pattern may reflect increased vegetation growth following wetter years. Such growth has been linked to greater fuel loads, a contributing factor in wildfire severity \citep{digiuseppe2025global}.

\section{Discussion}
\label{discussion}

Live Fuel Moisture Content (LFMC) is a key environmental variable for assessing wildfire risk, as it directly affects both the likelihood of ignition and the potential for fire spread. However, current live fuel moisture sampling is restricted by field and processing constraints, as samples must be collected from easily accessible areas and quickly transported to laboratory facilities for drying. These limitations reduce the spatial and temporal resolution of traditional measurements.

An accurate and reliable LFMC prediction model may offer significant value to land managers, wildland firefighters, and the communities potentially threatened by wildfires. Improved ability to analyze past wildfire events enhances our understanding of the conditions that contribute to large, destructive fires. This enables more effective anticipation and mitigation strategies. Furthermore, such insights support the evaluation of conditions under which prescribed fire has been successfully implemented, thereby promoting its strategic use as a tool for reducing wildfire risk.

This study demonstrates that a practical LFMC model can be built using existing state-of-the-art geospatial foundation models, achieving reasonably accurate predictions in many cases. The model offers high spatial resolution (10 meters), whereas previous studies provided only 500-meter resolution \citep{miller2023projecting}. A case study of 2025 wildfire events in Los Angeles supported the model's reliability, showing results consistent with observed surface conditions.

Nonetheless, the model has several limitations that warrant further research and development. While the Galileo-Tiny foundation model was trained on global data, the LFMC model was fine-tuned exclusively on data from the continental United States, primarily in the western region. A more extensive global sample could be acquired and used for training. Furthermore, the model has only been applied retrospectively and has not yet been tested for forecasting LFMC. Finally, it currently produces only monthly averages, whereas more frequent weekly and daily predictions---including on-demand generation for current conditions---would be highly valuable, especially for real-time wildfire management.

\section{Acknowledgments}

We thank John J. Battles (UC Berkeley), Rob York (UC Berkeley), Peter Stine (UC Davis; California Wildfire and Forest Resilience Task Force), Jonathan McDuffey (USFS), and Anand Padmanabhan (University of Illinois Urbana-Champaign) for their valuable feedback on this project.

\bibliography{papers}
\bibliographystyle{icml2025}

\end{document}